\documentclass[letterpaper]{article} 
\usepackage{aaai24}  
\usepackage{times}  
\usepackage{helvet}  
\usepackage{courier}  
\usepackage[hyphens]{url}  
\usepackage{graphicx} 
\usepackage{natbib}  
\usepackage{caption} 
\usepackage{algorithm}
\usepackage{algorithmic}

\usepackage{amsmath}
\usepackage{amssymb}
\usepackage{mathtools}
\usepackage{amsthm}
\usepackage{xcolor}

\usepackage{graphicx}
\usepackage{subfig}
\usepackage{booktabs} 
\usepackage{enumitem}
\usepackage{nicefrac}
\usepackage{subfig}
\usepackage{makecell}
\usepackage{tabularx}

\newcommand{\mscomplex}{MSComplexTask}
\newcommand{\dataset}{TaskLAMA}
\newcommand{\task}[1]{\mathsf{#1}}
\newcommand{\graph}[1]{\mathcal{#1}}
\newcommand{\nodes}[1]{\mathcal{#1}}
\newcommand{\edges}[1]{\mathcal{#1}}

\newtheorem{definition}{Definition}
\usepackage{newfloat}
\usepackage{listings}
\DeclareCaptionStyle{ruled}{labelfont=normalfont,labelsep=colon,strut=off} 
\floatstyle{ruled}
\newfloat{listing}{tb}{lst}{}
\floatname{listing}{Listing}
\nocopyright 

\title{TaskLAMA: Probing the Complex Task Understanding of Language Models}
\author{
    Quan Yuan$^*$, Mehran Kazemi$^*$, Xin Xu\thanks{Contributed equally.}, Isaac Noble, Vaiva Imbrasaite, Deepak Ramachandran\\
}
\affiliations{
    \texttt{\{yquan, mehrankazemi, xxujasmine, isaacn, vimbrasaite, ramachandrand\}@google.com}\\
    Google Research
}

\usepackage{bibentry}

\begin{document}

\maketitle

\begin{abstract}
Structured Complex Task Decomposition (SCTD) is the problem of breaking down a complex real-world task (such as \emph{planning a wedding}) into a directed acyclic graph over individual steps that contribute to achieving the task, with edges specifying temporal dependencies between them. SCTD is an important component of assistive planning tools, and a challenge for commonsense reasoning systems. We probe how accurately SCTD can be done with the knowledge extracted from Large Language Models (LLMs). We introduce a high-quality human-annotated dataset for this problem and novel metrics to fairly assess performance of LLMs against several baselines. Our experiments reveal that LLMs are able to decompose complex tasks into individual steps effectively, with a relative improvement of 15\% to 280\% over the best baseline. We also propose a number of approaches to further improve their performance, with a relative improvement of 7\% to 37\% over the base model. However, we find that LLMs still struggle to predict pairwise temporal dependencies, which reveals a gap in their understanding of complex tasks.
\end{abstract}

\section{Introduction}
In their daily lives, people are involved in executing multiple tasks of different temporal granularity in order to achieve their varied goals. This may range from simpler tasks such as \emph{washing a cup} which may take seconds or minutes to complete, to more complex tasks such as \emph{planning a wedding} which may take many weeks or months to complete. 

There is abundant evidence that consciously decomposing a complex task into smaller sub-tasks leads to more efficient and reliable execution. For example, \citet{kokkalis2013taskgenies} show that people tend to achieve tasks better and faster when given a concrete plan with actionable steps. \citet{cheng2015break} show that breaking a macro-task into micro-tasks for workers results in more accurate overall quality and allows for easier recovery from interruption. \citet{chilton2013cascade} show that breaking down a task into sub-tasks is beneficial for taxonomy creation. \citet{teevan2016supporting} show that breaking a task into sub-tasks can be beneficial for collaborative writing (e.g., writing a description of a shared project). \citet{allengetting} argue in their book that \emph{``there is an inverse relationship between things on your mind and those things getting done''}.

\begin{figure}[t]
  \centering
  \includegraphics[width=0.85\columnwidth]{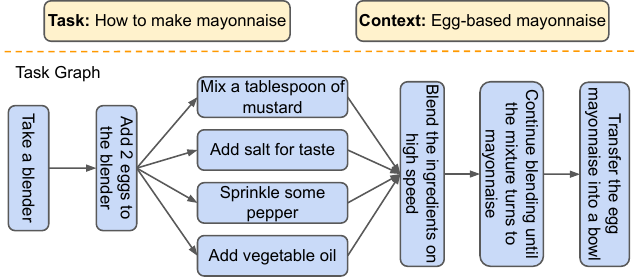}
  \caption{%
  \label{fig:dataset-example} %
  An example of a task graph for a complex task from the \emph{\dataset} dataset. Nine steps are outlined. The four steps with horizontal texts can be done in any order.
  }
\end{figure}

Given a complex task, the goal of structured complex task decomposition (SCTD) is to automatically find a complete set of necessary steps for achieving the task, and specify temporal dependencies between these steps (i.e. which steps should be done before which). The output can be described as a directed acyclic graph, which we term \emph{Task Graph}, where the nodes represent the steps and the edges represent temporal dependencies. Any ordering of the nodes that respects the edges of the graph is a possible ordering of the steps to accomplish the task. Figure~\ref{fig:dataset-example} demonstrates an example task graph for \emph{making egg-based mayonnaise}.  The importance of this problem has led to an extensive classic and modern literature for developing solutions based on the latest AI technologies available \cite{newell1972human,kokkalis2013taskgenies,hassan2014supporting,zhang2021learning}. These works were also motivated by the feature of SCTD being a quintessential or (informally speaking) AI-Complete problem, involving many features of human-level reasoning such as logical/defeasible reasoning, uncertainty and high context-dependence.
The existing solutions in the literature are typically based on crowd-sourcing \cite{kokkalis2013taskgenies,zhou2022show}, based on similarity or co-occurrence of user search queries \cite{hassan2014supporting,mehrotra2017extracting,zhang2015task}, or based on summarizing the content of relevant web-pages found through web search \cite{zhang2021learning}. 

In this paper, we probe the extent to which such knowledge can be extracted from large language models (LLMs). Previous work has shown that LLMs contain a large amount of different types of knowledge \cite{petroni2019language,sung2021can,west2021symbolic} and can do various types of reasoning \cite{wei2022chain,nye2021show,kazemi2023boardgameqa}. Our work extends this line of work by probing LLMs for their SCTD knowledge and reasoning abilities, and demonstrating their current strengths and limitations.

We create a high-quality human-annotated dataset for the SCTD problem. Following the current convention of naming language model probes as \emph{xLAMA}, we name the dataset \emph{\dataset}. Specifically, we gather a set of 1612 tasks and ask human annotators to 1) write their assumptions to provide context for the task, 2) write the required steps for those tasks under the provided context, and 2) specify the temporal dependencies between the steps. This gives us a total of 12118 steps and 11105 temporal dependencies.

We identify a potential problem with the metrics used in previous work to measure the quality of the generated nodes: one can arbitrarily improve the metric by simply adding duplicate sub-steps. To solve this issue, we propose robust metrics and report results with them providing the first fair comparison of methods for this problem. We also develop novel metrics for measuring the quality of the generated temporal dependencies. These metrics are potentially generally applicable in other settings where annotated/labeled graphs produced by generative models must be compared.

We compare the performance of LLMs on \dataset\ against heuristic-based, similarity-based, and query-based baselines. Our results reveal that the steps generated by an off-the-shelf LLM have higher quality than the baselines offering $15\%-280\%$ relative improvement compared to the best baseline in terms of different metrics. We also show that LLMs understand the context and can adapt the generated steps based on the context in which the complex task is to be done. We then propose a number of approaches to improve the performance of off-the-shelf LLMs even further, using the specialized structure of the SCTD problem. The combination of these solutions result in $7\%-37\%$ improvement over the base model, depending on the metric we use. We also measure the quality of the temporal dependencies produced by LLMs and observe that while LLMs are good at generating good sequences of steps, their ability in predicting pairwise temporal dependency remains unsatisfactory.

A summary of our main contributions follow: 1- we create \dataset, a high-quality probe specifically focused on complex real-world task understanding, 2- we develop metrics for measuring model performance for SCTD, 3- we propose various LLM-based approaches for SCTD and compare against a number of baselines that do not leverage LLMs, 4- we conduct a comprehensive set of experiments showing that LLMs perform well at decomposing a complex task into a sequence of steps, but their understanding of temporal dependencies between these steps remains unsatisfactory.

\section{Related Work}
We categorize the works from the literature that relate to our paper as follows:

\paragraph{Crowd-Sourced SCTD:} One line of work uses crowd-sourcing for obtaining the steps and their temporal order for complex tasks. \citet{kokkalis2013taskgenies} develop a framework where users can find information about various tasks: if a similar task (measured using natural language processing tools) already exists in their database, the information about that task is shown to the user; otherwise, the task is sent for crowd-sourcing. \citet{zhou2022show} combine the information from the WikiHow website\footnote{\url{https://www.wikihow.com}} to produce hierarchical task trees. While crowd-sourcing may lead to high-precision task graphs, it is costly and may suffer from low-recall as new task graphs cannot be built on-the-fly for novel tasks.

\paragraph{Query-based SCTD:} Another commonly used approach for SCTD is by leveraging user search queries. \citet{hassan2014supporting} create sessions from the search queries of a commercial search engine and propose steps for complex tasks by finding the queries that frequently co-occurred with the complex task in different sessions. \citet{mehrotra2017extracting} propose a hierarchical clustering approach for search queries where the queries higher in the hierarchy correspond to tasks and their children represent steps to those tasks. \citet{zhang2015task} map queries to demands using external knowledge and mine frequent demand patterns. In this work, we compare against a number of query-based approaches.

\paragraph{Summarization-based SCTD:} \citet{zhang2021learning} proposed a summarization-based approach to SCTD where for a given complex task, first a web search is done to identify relevant web pages, and then a language model is trained to summarize the contents of those web-pages into task graphs. In this work, we take a different approach by measuring how much of the information can be directly obtained from the LLM itself.

\paragraph{LLM Knowledge Probing:} Previous work has shown that LLMs contain a large amount of different types of knowledge. This includes
factual \citep{petroni2019language,jiang2020can}, commonsense \citep{zhou2020evaluating,davison2019commonsense,yin2022geomlama}, biomedical \citep{sung2021can}, numerical \citep{lin2020birds}, scale \citep{zhang2020language}, and many other types of knowledge. 
Most related to our work, it has been shown that LLMs perform well in breaking a simple goal into specific low-level actions a robot needs to take to achieve the goal \cite{huang2022language} (e.g., providing the low-level steps for a goal such as \emph{throw away garbage)}; they also perform well in simple, advice-seeking scripts \cite{sakaguchi2021proscript,madaan2022language,brahman2023plasma} (e.g., \emph{go out with friends}, \emph{live somewhere warmer}, etc.).
Our work is in the same vein with these works, but extends them by measuring the amount of information one can extract from LLMs for complex tasks requiring multiple (potentially complex) steps to be completed (see Table~\ref{tab:sample-tasks} for a sample of such tasks).

\begin{table}[t]
    \centering
    \footnotesize
    \caption{Sample complex tasks and extra assumptions (context) from \emph{\dataset}\ (-- means no extra assumption).}
    \begin{tabular}{c|c}
        \toprule
        Complex Task & Assumption / Context \\ \hline
        Build a curved retaining wall & Using concrete \\
        Start a property management company & In Florida \\
        Write a grant proposal & For non-profit \\
        Cook lobster tails at home & Grilled  \\
        Install a light switch & -- \\
        Get a real estate license & In Texas \\
        Recover deleted photos & From iPhone \\
        Plan a wedding & In Italy \\
        Become a travel agent & Online agent \\
        \bottomrule
    \end{tabular}
    \label{tab:sample-tasks}
\end{table}

\section{The \dataset\ Probe}
We create a dataset of task graphs for 1630 complex tasks. Following the \textbf{LA}nguage \textbf{M}odel \textbf{A}nalysis (LAMA) naming convention, we call our dataset \emph{TaskLAMA}. Before describing the dataset creation process, we start with defining our notation.
We represent a graph $\graph{G}=(\nodes{V}, \edges{E})$ as a tuple with $\nodes{V}=\{v_1, v_2, \dots, v_n\}$ representing the nodes and $\edges{E}\subset \nodes{V}^2$ represent edges. Throughout this paper, we work with directed graphs where, for an edge $(v_i, v_j)$, the order of the nodes is important.

A task graph is defined as follows:
\begin{definition}[Task Graph]
A task graph for a complex task $\task{T}$ is a graph $\graph{G}=(\nodes{V}, \edges{E})$ where each node $v_i\in\nodes{V}$ represents a step required for accomplishing $\task{T}$ and each edge $(v_i, v_j)\in\edges{E}$ represents a temporal dependence between $v_i$ and $v_j$, indicating that $v_i$ should be done before $v_j$.
\end{definition}

The problem we study in this paper is the following: Given a complex task $\task{T}$ (and sometimes an extra context about the task) as input, generate a task graph $\graph{G}$ as output. 
\dataset\ provides a probe for this problem with $(\task{T}_i, \graph{G}_i)$ pairs. 
The creation of \dataset\ involves four main components: 1- selecting a set $\{\task{T}_1, \dots, \task{T}_\tau\}$ of complex tasks, 2- gathering steps $\nodes{V}_i$ involved in the execution of these tasks, 3- gathering temporal dependencies $\edges{E}_i$ between the steps, and 4- splitting the dataset into train, validation, and test sets. In what follows, we explain each component in detail\footnote{The full dataset can be downloaded from \url{https://storage.googleapis.com/gresearch/tasklama/tasklama.zip}}.

\subsection{Selecting a set of complex tasks}
We obtain a varied and representative set of complex tasks performed by humans from the following two sources.

\textbf{The MSComplexTasks dataset} \cite{zhang2021learning}: There are $711$ distinct tasks in this dataset coming from the logs of Wunderlist, a popular task management application. These tasks are selected from a bigger pool of tasks using a number of filters, most notably filtering simple tasks. 

\textbf{Popular \emph{How To} search queries}: From the logs of a commercial search engine, we extracted popular search queries  that start with \emph{How To}. To ensure anonymity, we removed any query issued by fewer than 1000 unique users. We then deduplicated these tasks and applied a number of other filters to remove tasks involving sensitive and harmful topics, tasks requiring medical advice, or tasks that did not deem complex (e.g., tasks that did not involve multiple steps). We labeled the remaining tasks based on topic and sampled from each topic to avoid over-presence or under-presence from certain topics. At the end, we obtained $901$ tasks.

\subsection{Gathering steps for the tasks}
Complex tasks can be typically accomplished in multiple different ways, with a potentially different set of steps involved each time depending on the context (e.g., \emph{Make a burger} can have different steps depending on whether we do it \emph{On a charcoal grill} or \emph{In an air fryer}).
To gather the steps $\nodes{V}$ for a task $\task{T}$ while taking the context into account, we instructed annotators to do the following: for each task, write down the assumptions they are making and then the set of steps for the task under those assumptions. Some examples of tasks and assumptions are presented in Table~\ref{tab:sample-tasks}.

The annotators were allowed to search online and learn about the steps, but were required to then write the steps in their own words and based on their own understanding. The annotators were also instructed to make sure that each step starts with a verb, corresponds to exactly one action, is meaningful as a standalone sentence/does not contain anaphora (e.g., avoid \emph{put it on the grill}), is actionable as opposed to general advice, and is applicable in the context of the assumptions made. If the annotator was unfamiliar with the task, they were instructed to skip it; the task was then sent to another annotator.

We trained the annotators over three rounds of pilot study, each time having them annotate a small number of tasks and then explaining to them the mistakes they made. Some of the dominant mistakes in the initial rounds included directly copying search results, failing to follow one of the rules mentioned above, and/or misunderstanding how to provide assumptions. These issues were mostly resolved over the three rounds of training. In the final round, the workers spent an average of $892$ seconds on each task.
Through the above process, we gathered a total of 12118 steps for our 1612 tasks, with an average of $7.5$ steps per task.

\subsection{Gathering temporal dependencies}
Once we gathered the assumptions and steps for each task, a separate set of annotators were asked to specify the order dependencies $\edges{E}$ for the steps of each task. The annotators were instructed to first draw the graph on a piece of paper and then submit the edges one by one. Similar to the previous case, we trained the workers over three rounds of pilot studies. The mistakes in the initial rounds ranged from producing a linear sequence instead of a graph, only providing a partial graph (i.e. only a subset of the edges), and misunderstanding the concept of temporal dependence. These issues were mostly resolved over the three rounds of training. In the final round, the workers spent an average of $1138$ minutes on each task.
Through this process, we gathered a total $11105$ temporal dependencies for our 1612 tasks, with an average of $6.9$ dependencies per task.

\subsection{Dataset splitting}
We split the data into train, validation, and test sets in such a way that the tasks are conceptually different in the three sets. Towards this goal, we first grouped the 1612 tasks into $621$ clusters based on their textual similarity and then randomly split the clusters into train, validation, and test sets. This splitting strategy ensures some amount of difference in the tasks in each set. Following this splitting strategy, we ended up with $965$ examples in the training set, $169$ in the validation set, and $478$ in the test set.

\section{Method}
To generate task graphs for a given complex task, a model needs to 1) generate the steps, and 2) decide the order dependency between the steps.

\subsection{Generating the steps}
To generate the steps for a task, we experiment with the following strategies:

\textbf{In-Context Learning (ICL):} In ICL \cite{brown2020language}, one provides a few demonstrations each containing an input and the expected output, followed by the query for which the model has to generate the output. The model learns the relation between the input and the output in context, and uses that to generate an output for the provided query.

\textbf{Multiple Sequences (MultSeq):} We notice that when we generate multiple sequences of steps for a task using the ICL approach, the sequences sometimes have complementary steps between them. To leverage this intuition, we generate $k$ sequences of steps using the ICL approach, setting the decoding temperature to 0.5 to allow for diverse generations. Then we deduplicate the steps and combine the remaining steps to obtain the final set of steps. The deduplication procedure is explained in the Appendix.

\textbf{Sample and Filter (S\&F)}: We notice that when we generate multiple sequences of steps for a task using the ICL approach, some of the sequences have higher quality than the others. To leverage this, we first train a separate model that scores the sequences generated by the ICL approach. Then, we generate multiple sequences of steps and select the one with the highest score according to the trained model. To train a model that can score the generated sequences, we first generate $16$ sequences per task for the tasks in our training set, then we evaluate each of the generated sequences with respect to the golden sequence and obtain a single number indicating how good that sequence is. This gives us a dataset of (Task, Sequence of Steps, Score). We then train a model that given a task and a sequence of steps predicts the score.

\textbf{Soft-Prompt Tuning (SPT)}: In the case of ICL, the in-context demonstrations we provide as input get mapped to the corresponding token embeddings that are then fed into the LLM. Recently, it has been shown that instead of using a fixed set of token embeddings as the in-context demonstrations, one can learn those embeddings based on training data to enable better in-context examples. This technique is typically referred to as \emph{soft-prompt tuning} \cite{lester2021power}. 
    We learn the prompt embedding based on our training data and decide the size of the prompt based on performance on our validation set.

\textbf{MultSeq + S\&F}: We generate $k'$ sequences of steps using the ICL approach, then select and combine the top $k$ sequences ranked by the S\&F model.

\textbf{MultSeq + SPT}: We combine $k$ sequences from the SPT model instead of the ICL model.

\textbf{S\&F + SPT}: We use the S\&F model to score the sequences of steps generated by SPT and then select the best.

\textbf{MultSeq + S\&F + SPT}: This is similar to S\&F + SPT except that we combine the steps from the top $k$ sequences.

\subsection{Generating the order dependencies} \label{sec:link-approaches}
While task graphs are directed acyclic graphs (e.g., see Figure~\ref{fig:dataset-example}), when using an LLM to generate the steps for a task we get a sequence of steps. We compare a few approaches that can turn the sequence of steps generated by the LLM into a task graph.

\textbf{Linear:} We use the linear order of steps produced by the LLM as the final task graph.

\textbf{ICL:} We provide multiple examples as demonstrations each containing a task, two of its steps, and the label indicating whether the first step should be done before the second one. We then provide a new task and two of its steps and ask for the label. 

\textbf{ICL with Chain-of-Thought:} Chain-of-thought (CoT) prompting \cite{wei2022chain} is a technique where besides providing the input and the label, the demonstrations also provide a rationale for the label. We test a version of ICL with CoT where the rationale for the demonstrating examples are written manually.

\textbf{SPT:} We soft-prompt tune the LLM on the training data to learn to predict the label given a task and two of its steps.

\textbf{LLM Scoring:} Given the initial linear order produced by the LLM, we generate $m$ sequences by randomly swapping the order of two steps and use the LLM to score the sequences\footnote{Note that LLMs can be used both to generate an output and also to score a provided output.}. We then sort the sequences descendingly based on their LLM score and select the highest scoring sequences. Then, for two steps $v_i$ and $v_j$, if we see $v_i$ before $v_j$ in some sequences and $v_j$ before $v_i$ in the other sequences, we assume $v_i$ and $v_j$ can be done in any order; otherwise, if $v_i$ always appears before $v_j$ (or vice versa), we assume $v_i$ has to be done before $v_j$ (or vice versa). We turn the sequences into a graph following the above strategy.

In the case of cycles, i.e. if a model predicts that A should be done before B, B should be done before C, and C should be done before A, we remove the dependencies assuming that each of the steps can be done before the other one so they can be done in any order.

\begin{table*}[t]
    \centering
    \small
    \newcolumntype{C}{>{\centering\arraybackslash}X}
    \caption{The performance of different models for task step generation measured in terms of multiple metrics.}
    
    \begin{tabular*}{\textwidth}{m{0.18\textwidth}|>{\centering\arraybackslash}m{0.055\textwidth}|>{\centering\arraybackslash}m{0.055\textwidth}|>{\centering\arraybackslash}m{0.055\textwidth}|>{\centering\arraybackslash}m{0.055\textwidth}|>{\centering\arraybackslash}m{0.055\textwidth}|>{\centering\arraybackslash}m{0.055\textwidth}|>{\centering\arraybackslash}m{0.055\textwidth}|>{\centering\arraybackslash}m{0.055\textwidth}|>{\centering\arraybackslash}m{0.055\textwidth}|>{\centering\arraybackslash}m{0.055\textwidth}}
    
        \toprule
        - & \multicolumn{2}{c}{Rouge1} & \multicolumn{2}{c}{Rouge2} & \multicolumn{2}{c}{RougeL} & \multicolumn{2}{c}{Hungarian} & \multicolumn{2}{c}{Relaxed Hung.}  \\ \hline
        Model & F1 & F2 & F1 & F2 & F1 & F2 & F1 & F2 & F1 & F2 \\ \hline
        Repeat Task & 12.2 & 10.3 & 1.7 & 1.5 & 12.1 & 10.2 & 34.6 & 33.0 & 41.3 & 37.7   \\
        Repeat Sim & 11.6 & 10.2 & 1.5 & 1.3 & 10.8 & 9.5 & 33. & 33.4 & 40.1 & 38.4 \\
        Co-occur & 16.4 & 15.0 & 2.7 & 2.5 & 13.6 & 12.5 & 34.8 & 34.7 & 41.7 & 40.0 \\
        Hierarchical & 14.6 & 12.6 & 2.3 & 2.0 & 12.7 & 11.0 & 34.3 & 34.1 & 41.2 & 39.3 \\ \hline
        ICL & 33.1 & 31.7 & 10.0 & 9.5 & 24.0 & 23.0 & 40.1 & 40.3 & 48.1 & 47.3 \\
        MultSeq & 32.5 & 37.2 & 10.5 & 12.1 & 22.9 & 26.3 & 35.3 & 42.8 & 47.5 & 50.4 \\
        S\&F & 36.6 & 34.3 & 11.3 & 10.6 & 25.6 & 24.1 & 42.7 & 41.9 & 50.2 & 49.1 \\
        SPT & 38.7 & 36.3 & 13.2 & 12.3 & \textbf{26.7} & 25.1 & 43.6 & 42.4 & 51.5 & 50.3 \\ \hline
        MultSeq + S\&F & 37.6 & 38.3 & 12.1 & 12.3 & 25.2 & 25.7 & 41.7 & 44.4 & 50.5 & 51.1 \\
        MultSeq + SPT & 36.8 & \textbf{41.3} & 13.4 & \textbf{15.0} & 25.3 & \textbf{28.4} & 39.6 & \textbf{46.2} & 51.3 & \textbf{53.2} \\
        S\&F + SPT & \textbf{39.0} & 38.2 & 13.4 & 13.1 & \textbf{26.7} & 26.2 & \textbf{43.8} & 43.2 & 51.5 & 50.4 \\
        MultSeq + S\&F + SPT & 38.7 & 40.0 & \textbf{13.7} & 14.1 & 25.8 & 26.7 & 42.5 & 44.5 & \textbf{51.6} & 51.9 \\
        \bottomrule
    \end{tabular*}
    \label{tab:main-results}
\end{table*}

\section{Metrics}
For evaluation, we need two sets of metrics: one for measuring the quality of the steps (nodes) and one for measuring the quality of the temporal dependencies (edges). We discuss each of these separately.

\subsection{Node Metrics}
To compare the generated steps and measure their quality with respect to the golden steps, let $\nodes{V}_G=\{v_1, \dots, v_n\}$ be the steps in the golden graph, $\nodes{U}_M=\{u_1, \dots, u_m\}$ be the steps in the generated graph, and $S$ be a pairwise similarity function of the steps (we use the cosine similarity of the universal sentence encodings \cite{cer2018universal} of the steps).

Previous work has proposed to compute precision as $\sum_i \frac{\max_j S(u_i, v_j)}{|\nodes{U}_M|}$ and recall as $\sum_j \frac{\max_i S(u_j, v_i)}{|\nodes{V}_G|}$ \cite{zhang2021learning}. We find, however, that with these metrics, one can arbitrarily increase the precision without sacrificing recall. Consider the case where $\nodes{V}_G=\{v_1, v_2\}$, $\nodes{U}_{M_1} = \{u_1, u_2\}$, and $\nodes{U}_{M_2} = \{u_1, u'_1, u_2\}$ and let $S(u_1, v_1)=0.6$ and zero for other pairs, and $u'_1$ be a near-duplicate of $u_1$. Ideally, the output of $M_1$ should be preferred to the output of $M_2$ because they provide the same information but $M_1$ has no duplicates. However, using the above formulae, $M_1$ will have a precision of $\nicefrac{(0.6 + 0.0)}{2}=0.3$ and a recall of $\nicefrac{(0.6 + 0.0)}{2}=0.3$, whereas $M_2$ will have a precision of $\nicefrac{(0.6+0.6+0.0)}{3}=0.4$ and a recall of $\nicefrac{(0.6 + 0.0)}{2}=0.3$. We observed this issue in our experiments too, where combining the steps from multiple sequences (without deduplication) increased both precision and recall.

The problem described above is due to the one-to-many mapping formulation of precision and recall. To solve this issue, we use a Hungarian matching \cite{kuhn1955hungarian} of the steps that enforces a one-to-one mapping. We note, however, that in some cases, one step in the golden graph may correspond to more than one steps in the generated graph and vice versa, in which case a one-to-one mapping may be too restrictive. For example, the golden graph may contain two steps \emph{add salt} and \emph{add pepper} and the generated graph may contain a step \emph{add salt and pepper}. To account for such cases, we also report a relaxed version of Hungarian matching where we allow a one-to-two mapping\footnote{We could also report more relaxed versions such as one-to-three, but we observe that the cases where a single step corresponds to more than two steps are rare.}. 

Once the precision and recall are computed using (relaxed) Hungarian matching, we compute the $F1$ score and report it. We note that some of the generated steps that do not appear in the golden steps may still be good steps (e.g. because they provide more detail that is not in the golden graph). For this reason, recall might be a more informative metric than precision. To account for this, we also report an $F2$ score where we weigh recall twice as much as precision.

Following previous work \cite{zhang2021learning,madaan2022language}, we also concatenate the steps for each task and create a single document, and then report Rouge (F1 and F2) scores for these documents.

\subsection{Edge Metrics}
To compare the generated edges with those of the ground truth graph, we first match the nodes from the generated graph to the nodes from the golden graph using Hungarian matching. If a node in one graph does not have a match in the other graph, then we connect it to a dummy singleton node. Then, for each pair of matched nodes $(v_i, u_j)$, let $P_i$ and $C_i$ be the parents and children of $v_i$ respectively, and $P_j, C_j$ be the parents and children of $u_j$ respectively. We measure the amount of overlap between $P_i$ and $P_j$ as well as the amount of overlap between $C_i$ and $C_j$, both computed in terms of Rouge score. Then we report two metrics: 1- \textbf{In-Degree:} the average overlap between $P_i$ and $P_j$ over all matched pairs of steps, 2- \textbf{Out-Degree:} the average overlap between $C_i$ and $C_j$ over all matched pairs of steps. Intuitively, by measuring the overlap between $P_i$ and $P_j$, we measure the amount of overlap between the (immediate) preconditions of the two matched nodes. Moreover, by measuring the overlap between $C_i$ and $C_j$, we measure the amount of overlap between the steps that become executable (immediately) after we execute the matched nodes.

Furthermore, we also report another metric, which we term \textbf{step proximity}, computed as the average overlap between $P_i\cup C_i$ and $P_j \cup C_j$. Note that this metric does \emph{not} evaluate the order of the temporal dependencies; instead, it evaluates whether the steps that should be done in close proximity to each other are indeed placed close to each other in the generated graph.

\begin{figure*}[t]
  \centering
  \includegraphics[width=0.75\textwidth]{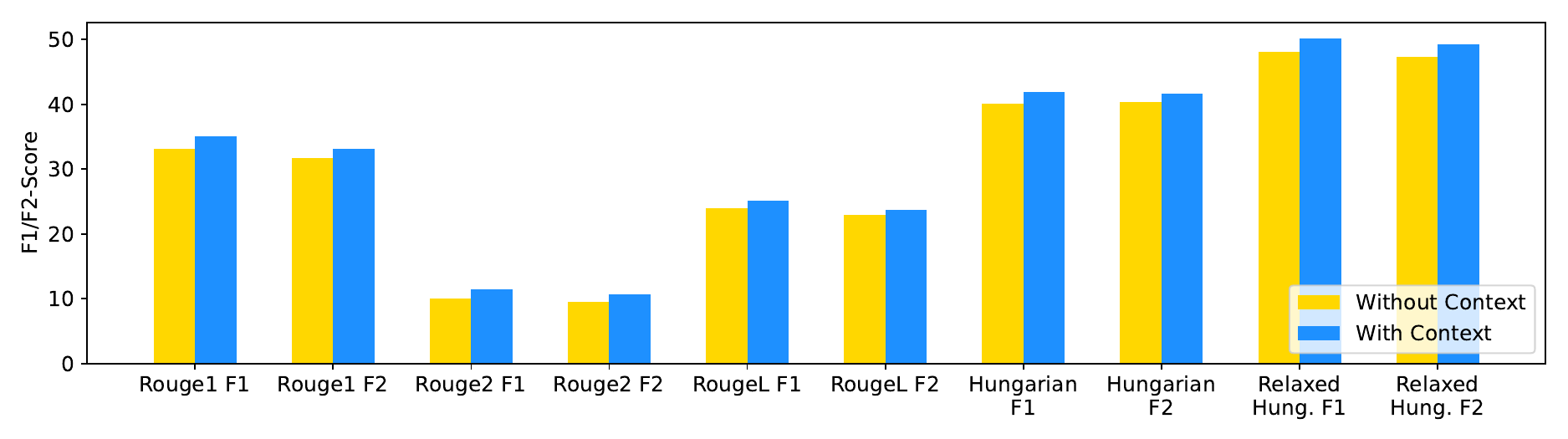}
  \caption{%
  \label{fig:context} %
  The model performance with and without the context provided as input. When the context is provided to the model, it performs better on all metrics.
  }
\end{figure*}

\section{Task Decomposition Results}

We compare against a number of baselines outlined below. 
    
\textbf{Repeat Task}: This baseline repeats the task $m$ times; $m$ is a hyperparameter that we tune on the validation set.

\textbf{Repeat Similar}: This baseline works similarly as the previous baseline, but for the $i$-th step, we randomly select one of the (non-stop word) tokens in the step and replace it with a semantically similar token. The similarity is computed based on GloVe embeddings \cite{pennington2014glove} of the tokens. The rationale for this substitution is that those words are likely to appear in the steps. For example, for the task \emph{make a smoothie}, some of the most similar words to \emph{smoothie} include \emph{yogurt}, \emph{juice}, \emph{strawberry} and \emph{granola} which are likely to appear in the steps of the task.

\textbf{Search Query Co-occurrence:} We cluster a large set of queries from a commercial search engine aiming at placing near-duplicate queries into the same cluster. Then, inspired by \citet{hassan2014supporting}, for a given task $\task{T}$ we find the cluster $\mathcal{C}_t$ that is most similar to $\task{T}$ and we rank the other clusters based on the co-occurrence of their queries with those of $\mathcal{C}_t$ and pick the top $k$ clusters. We select a representative query from each of these cluster to serve as a step for the task $\task{T}$. Here, $k$ is a hyperparameter that is tuned on the validation set.

\textbf{Search Query Hierarchy:} We perform a second level of clustering on top of the clustering from the previous baseline, where we aim to put similar-intent queries into the same cluster. Then, inspired by \citet{mehrotra2017extracting}, for a given task $\task{T}$ we find the clusters $\mathcal{C}_{L1}$ and $\mathcal{C}_{L2}$ from our level one and level two clustering (note that $\mathcal{C}_{L1}$ is a child of $\mathcal{C}_{L2}$) where $\task{T}$ should belong, and then obtain steps by selecting a query from each of the top $k$ sibling clusters of $\mathcal{C}_{L1}$. Here, $k$ is a hyperparameter that is tuned on the validation set.

\paragraph{Results:} The results are reported in Table~\ref{tab:main-results}. We observe that even the ICL approach significantly outperforms the other baselines on all metrics. For example, we observe a relative improvement of $280$\% over the best baseline in terms of Rouge2 F2-Score, and $15$\% in terms of Hungarian matching F1-score. This establishes LLMs as a powerful source for extracting information about the steps of a task.
Our solutions further improve upon the ICL results. In particular, we observe that both S\&F and SPT result in improvements compared to ICL across various metrics, with SPT providing more improvement compared to S\&F. The \emph{MultSeq} method brings improvement mostly for the F2-scores, due to having higher recall, showing that the steps generated in multiple LLM calls could be complementary as the combination of them improves recall, and hence the F2-score.

The approaches that mix two solutions perform better than the individual approaches in isolation in many cases. Among these approaches, we observe that \emph{MultSeq + SPT} works best in terms of the F2-score and \emph{S\&F+SPT} performs best in terms of F1-score. The combination of all three solutions also works well, but is often dominated by one of the approaches that combines two solutions.

\begin{figure}[t]
  \centering
  \includegraphics[width=0.8\columnwidth]{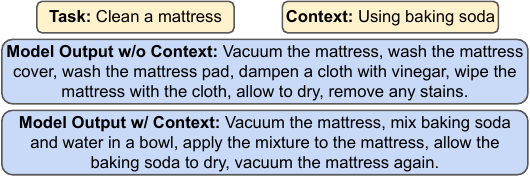}
  \caption{%
  \label{fig:context-example} %
  An example of model outputs for a task with and without the context provided as input to the model.
  }
\end{figure}

\begin{table*}[t]
    \centering
    \small
    \newcolumntype{C}{>{\centering\arraybackslash}X}
    \caption{The performance of different models for edge evaluation on the generated graphs (\#seqs=1 corresponds to the SPT model and \#seqs=2 to the SPT+MultSeq model).}
    
    \begin{tabular*}{0.87\textwidth}{m{0.10\textwidth}|>{\centering\arraybackslash}m{0.05\textwidth}|>{\centering\arraybackslash}m{0.05\textwidth}|>{\centering\arraybackslash}m{0.05\textwidth}|>{\centering\arraybackslash}m{0.05\textwidth}|>{\centering\arraybackslash}m{0.05\textwidth}|>{\centering\arraybackslash}m{0.05\textwidth}|>{\centering\arraybackslash}m{0.05\textwidth}||>{\centering\arraybackslash}m{0.05\textwidth}|>{\centering\arraybackslash}m{0.05\textwidth}|>{\centering\arraybackslash}m{0.05\textwidth}}
    
        \toprule
        - & - & \multicolumn{3}{c}{In-Degree} & \multicolumn{3}{c}{Out-Degree} & \multicolumn{3}{c}{Step Proximity}  \\ \hline
        Model & \#Seqs & Rouge1 & Rouge2 & RougeL & Rouge1 & Rouge2 & RougeL & Rouge1 & Rouge2 & RougeL \\ \hline
        Linear Order & 1 & 18.3 & 8.8 & \textbf{17.7} & 17.3 & 7.5 & \textbf{16.7} & \textbf{20.2} & \textbf{5.9} & \textbf{18.2} \\
        ICL & 1 & 13.6 & 8.0 & 13.3 & 13.6 & 7.6 & 13.3 & 15.2 & 4.0 & 13.8 \\
        ICL with CoT & 1 & 14.0 & 6.4 & 13.5 & 13.7 & 5.9 & 13.2 & 18.2 & 4.8 & 16.2\\
        SPT & 1 & 18.0 & \textbf{8.9} & 17.3 & \textbf{17.4} & \textbf{8.2} & \textbf{16.7} & 19.9 & 5.6 & 17.8 \\ 
        SPT + Linear & 1 & 18.1 & \textbf{8.9} & 17.4 & 17.1 & 8.1 & 16.4 & 20.0 & 5.7 & 17.9 \\
        LLM Scoring & 1 & \textbf{18.4} & 8.8 & \textbf{17.7} & 17.2 & 7.5 & 16.6 & \textbf{20.2} & \textbf{5.9} & \textbf{18.2}\\
        \hline \hline
        ICL & 2 & 10.3 & 4.4 & 10.0 & 10.1 & 4.2 & 9.8 & 13.6 & 3.5 & 12.3 \\
        ICL with CoT & 2 & 10.8 & 3.8 & 10.3 & 10.5 & 3.9 & 10.0 & 14.4 & 3.7 & 12.8 \\
        SPT & 2 & \textbf{13.1} & \textbf{5.8} & \textbf{12.7} & \textbf{12.8} & \textbf{5.5} &  \textbf{12.3} & \textbf{15.1} & \textbf{4.2} & \textbf{13.5}\\
        \bottomrule
    \end{tabular*}
    \label{tab:edge-eval-generated}
\end{table*}

\begin{table}[t]
    \centering
    \footnotesize
    \caption{The performance of different models for edge evaluation on the golden graphs. The \emph{Majority Class} baseline always predicts no dependency.}
    
    \begin{tabular}{c|c}
        \toprule
        Model & Accuracy \\ \hline
        Majority Class & 53.8 \\
        ICL & 47.5 \\
        ICL with CoT & 49.6 \\
        SPT & 78.6 \\ 
        \bottomrule
    \end{tabular}
    \label{tab:edge-eval-gold}
\end{table}

\subsection{Context Understanding Results}
The steps for completing a complex task can be different depending on the context. For example, \emph{recover deleted photos} has different steps depending on the device. We measure the ability of LLMs in providing contextualized steps for a task.

Recall that in our dataset, the steps for a task are written under certain assumptions that provide the context. To measure how well LLMs can providing contextualized steps for a task, we compare their performance with and without the assumption/context being provided to them. We test both settings with ICL: in one case we only provide the task and the steps and in the other we provide the task, assumptions/context, and the steps.

According to Figure~\ref{fig:context}, when we provide extra context, the model performs better across all metrics. This shows that LLMs are able to adjust the steps based on the context for a task.
In Figure~\ref{fig:context-example}, we provide an example model output with and without the context provided to the model. We can see that when no context is provided to the model, the model either provides general steps or selects a specific approach (using vinegar), but when the context is provided the model provides steps that are more specific to the context.

\section{Temporal Dependency Results}
We next verify how well LLMs can predict the temporal dependencies between the steps of the tasks (i.e. the edges of the task graphs). Initially, we compare the ICL, ICL+CoT, and SPT approaches on the golden nodes and edges (note that the other two edge prediction approaches are only applicable to generated graphs). In this case, for each pair of nodes in each of the golden task graphs in the test we ask the model to predict if one step should be done before the other one and report the accuracy. The results are reported in Table~\ref{tab:edge-eval-gold}. According to the results, both ICL and ICL+CoT perform quite poorly, but the performance improves massively after soft-prompt tuning, thus showing that temporal dependency understanding does not surface out of the box from LLMs but requires tuning on some data.

We next evaluate the performance of the approaches on the generated graphs. To this end, we conduct two experiments in one case we fix the generated steps to those of the SPT model (i.e. there is only one sequence of steps) and in the second we fix the generated steps to those of the SPT+MultSeq model (i.e. there are multiple sequences of steps). We then use the aforementioned approaches to decide the links. 
The results are reported in Table~\ref{tab:edge-eval-generated}. According to the results, in the case where we use only one sequence, we observe that the linear order produced by the LLM is quite a strong baseline: it outperforms the other models for step proximity (except for LLM Scoring where the two models are on-par) and only slightly underperforms in terms of in-degree and out-degree metrics. We also tried a version of the SPT where we combine it with the linear order of the LLM by making the following assumption: if the LLM produced step A before step B, then we assume either A should be done before B, or A and B can be done in any order (i.e. we rule out the possibility that B should be done before A). Even in this case, we observe that the linear model alone still gives a better performance. Our results show that while LLMs are good at generating the sequence of steps for a task in the right order, they are not particularly good at individually deciding which step of a task should be done before the other or if two steps can be done in any order.

\begin{figure}[t]
  \centering
  \includegraphics[width=0.9\columnwidth]{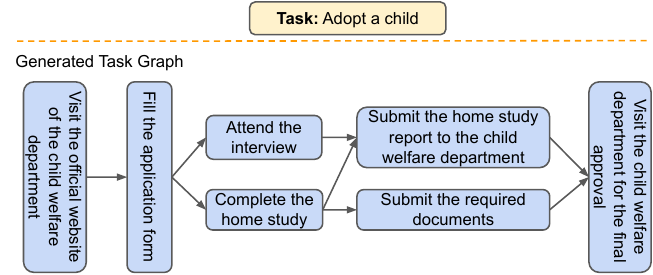}
  \caption{%
  \label{fig:gen-graph-adopt} %
  An example of an LLM-generated task graph.
  }
\end{figure}

\section{Qualitative Analysis}
In Figure~\ref{fig:gen-graph-adopt}, we demonstrate an LLM-generated task graph. We can see that many of the steps and the temporal dependencies are sensible; however, there may also be some problems present. For example, one probably needs to complete the home study before attending the interview. More qualitative examples are provided in the Appendix.

\section{Conclusion}
We studied the power of large language models (LLMs) for structured complex task decomposition (SCTD), i.e. the problem of decomposing a complex real-world task into multiple steps and determining the temporal dependencies between the steps. To this end, we created a probe named \dataset, developed metrics for measuring the performance, tested various task decomposition and temporal dependency prediction models and compared against baselines. Our results indicate that LLMs are strong in task decomposition. For predicting temporal dependencies, however, while they are able to produce a good sequence of steps with the right order of steps, their ability in predicting pairwise temporal dependencies still lags behind. Future work can find ways of improving the temporal dependency understanding of LLMs, or develop new approaches to improve LLM-based SCTD, e.g. by generating the entire task graph at once (see \citet{sakaguchi2021proscript,madaan2022language}) or by recursively breaking the complex tasks into simpler tasks and then solving those simpler task (see \citet{zhou2022least,kazemi2023lambada}). 

\bibliography{aaai24}


\appendix

\begin{table*}[t]
    \centering
    \tiny
    \newcolumntype{C}{>{\centering\arraybackslash}X}
    \caption{Quality issues in the \mscomplex\ dataset.
    }
    \begin{tabular*}{\textwidth}{m{0.10\textwidth}|>{\centering\arraybackslash}m{0.26\textwidth}|>{\centering\arraybackslash}m{0.33\textwidth}|>{\centering\arraybackslash}m{0.24\textwidth}}
        \toprule
        \textbf{Issue} & \textbf{Example 1} & \textbf{Example 2} & \textbf{Example 3} \\ \midrule
        
        Train/Test Overlap & \makecell{Test: build a wooden privacy fence \\ Train: build a wood privacy fence} & \makecell{Test: build an above ground pool deck \\ Train: Build a deck for an above ground pool} & \makecell{Test: Set up an etsy shop \\ Train: Start an etsy business} \\ \hline 
        
        Irrelevant Steps & \makecell{Task: Apply for citizenship uscis \\ Step: View a larger version of the infographic} & \makecell{Task: Paint front door \\ Step: Pin this to Pinterest} &  \makecell{Task: Apply for fmla \\ Step: View the archived webinar} \\ \hline
        
        Parsing issues & \makecell{Task: Apply for social security card \\ Step: Allowed To Work(See Instructions On Page 3)} & \makecell{Task: Register a business in Texas \\ Step: Doing Business As (DBA} &  \makecell{Task: Learn to hack \\ Step: Step-3: Learn Programming} \\ \hline
        
        Duplicate steps & \makecell{Task: Paint front door \\ Step: Choosing a color \\ Step: Choose New Paint Color} & \makecell{Task: Add music to google slideshow \\ Step: Add a YouTube Video \\ Step Add music from a YouTube video \\ Step: Method 2:- Add Music From YouTube} & \makecell{Task: fix a garage door \\ Step: Close the door \\ Step: Close the garage door}  \\ \hline
        
        Pronouns & \makecell{Task: Sell artwork \\ Step: Leave room for them to ask questions} & \makecell{Task: Apply for social security card \\ Step: Use this form to apply for a new or replacemet SSN card} & \makecell{Task: Build a murphy bed cheap \\ Step: Attach them with 2\u201d screws} \\ \hline
        (Irrelevant) Advice instead of Action & \makecell{Task: Sell artwork \\ Step: Change your art practice} & \makecell{Task: Write a job description \\ Step: Use bullet points} & \makecell{Task: Apply for fmla \\ Step: Work for a covered employer} \\
        \bottomrule
    \end{tabular*}
    \label{tab:mscomplex-issues}
\end{table*}

\begin{table*}[t]
    \centering
    \small
    \newcolumntype{C}{>{\centering\arraybackslash}X}
    \caption{The precision and recall of different models for task step generation measured in terms of multiple metrics.}
    
    \begin{tabular*}{\textwidth}{m{0.19\textwidth}|>{\centering\arraybackslash}m{0.054\textwidth}|>{\centering\arraybackslash}m{0.054\textwidth}|>{\centering\arraybackslash}m{0.054\textwidth}|>{\centering\arraybackslash}m{0.054\textwidth}|>{\centering\arraybackslash}m{0.054\textwidth}|>{\centering\arraybackslash}m{0.054\textwidth}|>{\centering\arraybackslash}m{0.054\textwidth}|>{\centering\arraybackslash}m{0.054\textwidth}|>{\centering\arraybackslash}m{0.054\textwidth}|>{\centering\arraybackslash}m{0.054\textwidth}}
    
        \toprule
        - & \multicolumn{2}{c}{Rouge1} & \multicolumn{2}{c}{Rouge2} & \multicolumn{2}{c}{RougeL} & \multicolumn{2}{c}{Hungarian} & \multicolumn{2}{c}{Relaxed Hung.}  \\ \hline
        Model & Prec. & Recall & Prec. & Recall & Prec. & Recall & Prec. & Recall & Prec. & Recall \\ \hline
        Repeat Task & 20.0 & 9.4 & 2.7 & 1.3 & 19.9 & 9.3 & 38.7 & 32.2 & 49.5 & 35.7 \\
        Repeat Sim & 1.7 & 9.5 & 2.0 & 1.2 & 15.7 & 8.9 & 34.8 & 33.5 & 43.6 & 37.4 \\
        Co-occur & 20.9 & 14.4 & 3.4 & 2.4 & 17.4 & 12.0 & 36.1 & 34.9 & 45.4 & 39.0 \\
        Hierarchical & 21.5 & 11.7 & 3.4 & 1.9 & 18.7 & 10.2 & 35.5 & 34.2 & 45.2 & 38.2 \\ \hline
        ICL & 39.4 & 31.4 & 12.0 & 9.4 & 28.6 & 22.7 & 41.9 & 40.9 & 49.9 & 46.9 \\
        MultSeq & 28.2 & 42.4 & 9.0 & 13.9 & 19.8 & 30.1 & 28.0 & 51.0 & 44.0 & 52.8 \\
        S\&F & 43.5 & 33.3 & 13.6 & 10.3 & 30.4 & 23.4 & 45.7 & 41.7 & 52.6 & 48.4 \\
        SPT & 46.5 & 35.1 & 15.9 & 11.9 & 32.1 & 24.3 & 47.8 & 41.9 & 53.8 & 49.7 \\ \hline
        MultSeq + S\&F & 39.0 & 39.4 & 12.7 & 12.6 & 26.3 & 26.5 & 39.5 & 47.2 & 50.0 & 51.6 \\
        MultSeq + SPT & 32.5 & 45.8 & 11.8 & 16.7 & 22.2 & 31.6 & 32.9 & 52.8 & 48.9 & 54.8 \\
        S\&F + SPT & 42.6 & 38.1 & 14.6 & 13.0 & 29.1 & 26.2 & 46.6 & 43.2 & 53.6 & 49.8 \\
        MultSeq + S\&F + SPT & 39.8 & 41.7 & 14.4 & 14.7 & 26.6 & 27.8 & 41.8 & 46.8 & 51.5 & 52.3 \\
        \bottomrule
    \end{tabular*}
    \label{tab:main-results-prec-recall}
\end{table*}

\section{The quality of existing datasets} \label{sec:appendix}
\mscomplex\ \cite{zhang2021learning} is the only dataset we are aware of that includes steps and temporal dependencies for complex tasks. However, we found several quality issues in the dataset that motivated us to create a new probe. Here, we categorize some of the main quality issues and provide a few examples for each category in Table~\ref{tab:mscomplex-issues}.
\begin{itemize}[topsep=0.2pt,itemsep=0.2pt,leftmargin=5mm]
    \item \textbf{Train/Test Overlap:} We found that for a large number of tasks in the test set, there exists one or more near duplicate (or closely related) task in the training set. To quantify the extent of this issue, for each task in the test set we first identified a set of highly similar tasks from the training set with a combination of manual search and automatic textual similarity. Then, we showed pairs of tasks to a human annotator who judged if the two tasks are 1- near duplicate (the two tasks being identical despite differences in the textual description of the task), 2- closely related (the two tasks are not identical but may share several sub-steps), or 3- other (the two tasks may not share several sub-steps). We found that for $21\%$ of the tasks in the test set, there was at least one near duplicate task in the training set, and for another $13\%$ there was a closely related task in the training set, amounting to $34\%$ leakage in total. 
    \item \textbf{Irrelevant steps:} For many tasks, we found that some of the provided steps are irrelevant to the task. 
    \item \textbf{Parsing issues:} Since the initial steps have been mined automatically from the web, we found several parsing issues resulting in redundant text, incomplete steps, extra information with respect to the order (potentially leading to leakage for determining order dependency), etc. 
    \item \textbf{Duplicate steps:} Since the initial steps come from multiple sources, we found that sometimes a task has several duplicated steps.
    \item \textbf{Coreferences:} Many steps contain pronouns that may refer either to something in the other steps or to something that is not even mentioned in the other steps, making the steps not standalone. 
    \item \textbf{(Irrelevant) advice instead of action:} Some of the steps provided for a task do not correspond to an action that the user has to take, but rather to a (sometimes irrelevant) advice.
\end{itemize}

Another related dataset is the ProScript \cite{sakaguchi2021proscript}. This dataset contains both steps and temporal dependencies. However, we find that the inputs are mostly simple or advice-seeking scripts where one of the many options can be selected. Examples include \emph{take a shower after work}, \emph{try daring foods}, \emph{see the forest}, \emph{live somewhere warmer}, \emph{go to a bar one day}, etc. We also find that many steps for the script are based on imaginary situations (e.g., \emph{take the elevator downstairs} is a step for the script \emph{meet for lunch}, which might be based on an imaginary situation where the person needs to take the elevator and go downstairs to meet for lunch).

\begin{figure}[t]
  \centering
  \includegraphics[width=0.9\columnwidth]{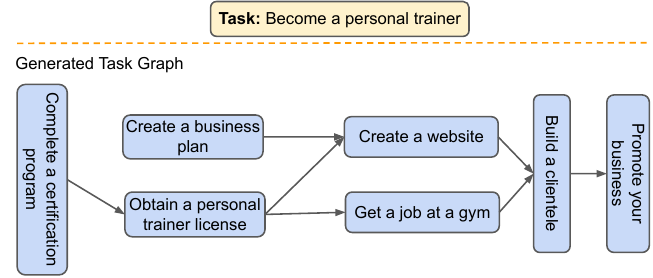}
  \caption{%
  \label{fig:gen-graph-trainer} %
  An example of an LLM-generated task graph.
  }
\end{figure}

\begin{figure}[t]
  \centering
  \includegraphics[width=0.9\columnwidth]{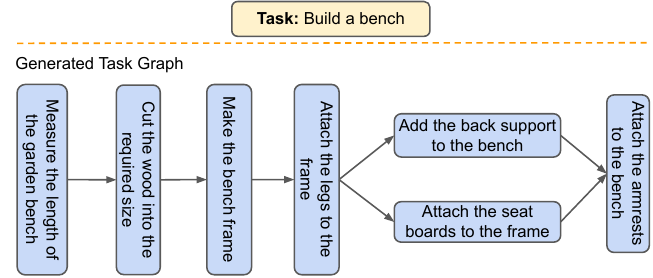}
  \caption{%
  \label{fig:gen-graph-bench} %
  An example of an LLM-generated task graph.
  }
\end{figure}

\begin{figure}[t]
  \centering
  \includegraphics[width=0.9\columnwidth]{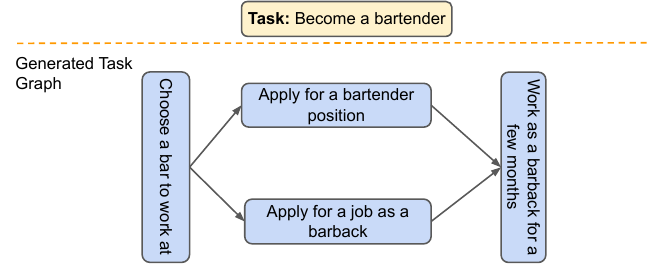}
  \caption{%
  \label{fig:gen-graph-bartender} %
  An example of an LLM-generated task graph.
  }
\end{figure}

\section{More Results}
\textbf{More qualitative examples:} 
Figures~\ref{fig:gen-graph-trainer}~and~\ref{fig:gen-graph-bench} show examples of LLM-generated task graphs where the nodes and edges are mostly meaningful. One possible edge mistake in Figures~\ref{fig:gen-graph-trainer} could be the one between \emph{Build a clientele} and \emph{Promote your business}, where one may expect to first promote the business and then build a clientele. Figures~\ref{fig:gen-graph-bartender} shows an example of a lower-quality LLM-generated task graph, where the steps are very generic and do not contain much information, and there are steps for applying for both bartender and barback positions, whereas the task is to become a bartender.

\textbf{Precision and Recall:} In the main text, for complex task decomposition we reported the F1 and F2 results. For completeness sake, in Table~\ref{tab:main-results-prec-recall} we report the precision and recall results. Similar conclusions can be derived as before with the S\&F and SFT models helping improve both precision and recall and the MultSeq model mostly helping recall.

\section{Implementation Details}\label{sec:implementation-details}
The experiments were done with the PaLM 62B model \cite{chowdhery2022palm} on TPUs of version 4. Due to the high cost of the experiments, the reported results are for one run only. 

\subsection{Task Decomposition}
In the case of the \emph{Repeat Similar} baseline, we sampled from the top-20 most similar tokens to that of the token that was to be replaced. In the case of ICL, we provided $10$ demonstrating examples from the training set as our in-context demonstrations. We select these $10$ examples randomly and leave more sophisticated selection approaches, e.g. see \cite{luo2023dr} as future work. We used the following prompt:\\
\texttt{EXAMPLE 1}\\
\texttt{What are the steps to [DEMONSTRATING TASK 1]? STEP1 [S1], STEP2 [S2], ..., STEPn [Sn].}
\texttt{...}\\
\texttt{EXAMPLE 10}\\
\texttt{What are the steps to [DEMONSTRATING TASK 10]? STEP1 [S1'], STEP2 [S2'], ..., STEPm [Sm'].}\\
\texttt{EXAMPLE 11}\\
\texttt{What are the steps to [TEST TASK]?}

In the case where we also provide an additional context, each example is as follows:\\
\texttt{What are the steps to [DEMONSTRATING TASK k] in the following context: [CONTEXT k]? STEP1 [S1], STEP2 [S2], ..., STEPn [Sn].}

For the soft-prompt tuned model, we tested prompt sizes from $\{100, 50, 10\}$ and selected the one that performed best on the validation set. We set the learning rate to $0.1$, batch size to $8$, and total number of training steps to $10000$. 

For the S\&F model, we trained a BERT-base model \cite{devlin2018bert} for sequence scoring with a learning rate of $1e-5$ and batch size of $16$, over $10$ epochs. We also tried BERT-large and BERT-small; while BERT-base was slightly better than BERT-small, the difference between BERT-base and BERT-large was negligible. 

For deduplication in the \emph{MultSeq} model, we embed the steps into a vector representation using the universal sentence encoding \cite{cer2018universal} and then consider two steps as being near-duplicates if the cosine similarity of their embedding is higher than a threshold. To decide the threshold, we generated multiple sequences of steps for the tasks in our training set, manually labeled some pairs of steps as being near-duplicates or being different, and then we selected a threshold that maximized the classification accuracy on the labeled set.

\subsection{Temporal Dependencies}
In the case of ICL, we provided $10$ demonstrating examples using the following prompt:\\
\texttt{EXAMPLE 1}\\
\texttt{For [DEMONSTRATING TASK 1], should I do sub-step "[Si]" before sub-step "[Sj]"? [Yes/No].}\\
\texttt{...}\\
\texttt{EXAMPLE 10}\\
\texttt{For [DEMONSTRATING TASK 10], should I do sub-step "[Si']" before sub-step "[Sj']"? [Yes/No].}\\
\texttt{EXAMPLE 11}\\
\texttt{For [TEST TASK], should I do sub-step "[Si'']" before sub-step "[Sj'']"?}

In the case where we add chain-of-thought, each example is as follows:\\
\texttt{For [DEMONSTRATING TASK k], should I do sub-step "[Si]" before sub-step "[Sj]"? [Rationale] therefore [Yes/No].}

For the soft-prompt tuned model, we set the prompt size to $10$, the learning rate to $0.1$, batch size to $4$, and the total number of training steps to $14000$.
For the \emph{LLM Scoring} model, we generate sequences by considering all possible ways of swapping two steps (truncated at 128 sequences at most), rank them based on their LLM score and filter out the bottom half, and construct the graph based on the remaining sequences.

\section{Limitations}
Measuring the complex task understanding of models is a challenging problem both in terms of dataset construction and evaluation. While we took a step toward this goal, we outline several limitations of our work that future work can resolve:
\begin{itemize}
\item \textbf{Conditional Task Graphs:} While \dataset\ provides unconditional task graphs, the steps (and order dependencies) in complex real-world tasks are typically conditional. As an example, after completing some the steps, the following steps might be different depending on the outcome of the previous steps. Future work can develop probes and benchmark models for conditional task graph generation.
\item \textbf{Granularity:} One important challenge for developing models (or measuring their performance) for task understanding is that Task Graphs can be generated at different levels granularity, making evaluation difficult. In this work, we only generated Task Graph at one level of granularity. Future work can find ways of generating and evaluating task graphs at different levels of granularity.
\item \textbf{Similarity vs entailment:} For evaluation, following previous work we measured step overlap in terms of similarity. However, a similarity-based approach fails to capture cases where two steps are not textually similar but one entails the other. Future work can use entailment models in place of similarity models for evaluation.
\item \textbf{Simultaneously generating steps and temporal dependencies:} In this work, we only examined the approaches where the steps (nodes) and temporal dependencies (edges) are produced independently. Future work can measure the performance of the models that produce both of them simultaneously.
\end{itemize}

\end{document}